\pdfoutput=1

\documentclass[11pt]{article}

\usepackage{acl}

\usepackage{times}
\usepackage{latexsym}

\usepackage[T1]{fontenc}

\usepackage[utf8]{inputenc}

\usepackage{microtype}

\usepackage{graphicx}

\usepackage{multirow}

\usepackage[T1]{tipa}
\usepackage{makecell}

\usepackage{url}

\usepackage{tablefootnote}

\usepackage{adjustbox}

\usepackage{url}

\usepackage{breakurl}
\usepackage[breaklinks]{hyperref}
\title{Suum Cuique: Studying Bias in Taboo Detection with a Community Perspective$^\star{}$}

\author{Osama Khalid$^\dagger$, Jonathan Rusert$^\dagger$, Padmini Srinivasan \\
  University of Iowa\\
  \texttt{\{osama-khalid, jonathan-rusert, padmini-srinivasan\}@uiowa.edu} \\}

\begin{document}

\maketitle

\begin{abstract}
  \vspace{-0.5em}
Prior research has discussed and illustrated the need to consider linguistic norms at the community level when studying \textit{taboo} (hateful/offensive/toxic etc.) language.
However, a methodology for doing so, that is firmly founded on community language norms is still largely absent.
This can lead both to biases in taboo text classification and limitations in our understanding of the causes of bias.
    We propose a method to study bias in taboo classification and annotation where a community perspective is front and center. This is accomplished by using special classifiers tuned for each community's language. In essence, these classifiers represent community level language norms.
    We use these to study bias and find, for example, biases are largest against African Americans (7/10 datasets and all 3 classifiers examined).
    In contrast to previous papers we also study other communities and find, for example, strong biases against South Asians.
    In a small scale user study we illustrate our key idea which is that common utterances, i.e., those with high alignment scores with a community (community classifier confidence scores) are unlikely to be regarded taboo.
    Annotators who are community members contradict taboo classification decisions and annotations in a majority of instances.
    This paper is a significant step toward reducing false positive taboo decisions that over time harm minority communities.
    
\end{abstract}

\thanks{\small $\star$~~This paper examines taboo language as a case study. The reader is cautioned that the paper contains strong language.}

\thanks{\small $\dagger$~~Equal Contribution.}


\section{Introduction}

Members of a community rely on a shared language for communication, one which evolves naturally, \textit{in situ}, and is shaped by the norms and mores of the community \cite{gumperz1968speech}.
Norms promote \textit{Entitativity}, 
the perception of group identity 
\cite{allan2006forbidden}.
Norms may be explicit and codified into law or so subtle that while members may be unable to specify them they can still recognize violations \cite{chandrasekharan2018internet}.
Utterances violating a these norms may be considered \textit{taboo} from that community's perspective.\footnote{We include hate, offense, sexism, toxicity, and abusive utterances in \textit{taboo} while acknowledging their nuanced differences \cite{fortuna2020toxic}.}  
%
%
 
%
Communities often self-regulate by censuring norm violating \textit{taboo} language 
\cite{allan2006forbidden}.  This censuring may be subtle such as ignoring taboo utterances leaving the individual somewhat isolated and ineffective. 
Self-regulation could also be explicit and even severe such as expulsion from the community.

\noindent \textbf{Taboos are community-specific:}
Perceptions of taboos in language use are influenced by community 
\cite{allan2006forbidden}.
Utterances that are benign in one community might be taboo in another\footnote{Even individuals within a community may differ in their norms and taboos, with some enforcing more constraints.}. E.g., `autistic' is considered derogatory on most of  
reddit, its use usually leads to censure. However, on the subreddit /r/wallstreetbets, it is not and is instead used as a self-descriptor by members.
The importance of considering the author's community when studying taboo is emphasized in a recent critical survey of bias in NLP papers \cite{blodgett-etal-2020-language}; it urges us to understand how social hierarchy relates to language use.  Our research is a step in this direction.

\noindent
\textbf{Taboo utterance detection and biases:}
Several papers especially from OffensEval 
\cite{zampieri2019semeval, zampieri2020semeval} 
have led to the development of state of the art taboo classifiers for moderating online speech. 
%
%
Alongside, an active research stream studies the presence of bias in taboo detection. 
Biases, specifically having higher false positive rates for a minority community compared to the majority, have been detected particularly against African Americans and also to some extent women \cite{dixon2018measuring,zhou2021challenges,bolukbasi2016man,xia-etal-2020-demoting,chuang2021mitigating}.
Bias is clearly harmful as penalties over false positives could over time discourage/constrain minority participation on social media platforms, and in turn reinforce the majority's norms and values.

%
%

\noindent
\textbf{Limitations in bias assessment research:}
A limitation is that community language norms are rarely considered.
While recent research on taboo detection discuss and illustrate the importance of community perspective  \cite{davidson2017automated, sap2019risk, badjatiya2019stereotypical} we do not yet have bias detection methods that are firmly founded on community language norms.
This methodological gap underlies not only the design of state-of-the-art taboo detection tools but perhaps more crucially even in the methodologies for the \emph{study} of bias in taboo detection. 
Not only are the taboo annotated datasets largely devoid of community, culture or social contexts (i.e., characterization)
\cite{zampieri2019semeval, zampieri2020semeval}, the study of bias itself needs strengthening with methods wherein a community perspective stays front and center, consistent with the urging in the survey \cite{blodgett-etal-2020-language}.

\noindent \textbf{Contributions:} 


\noindent 1. We propose a new method that is firmly grounded on community-specific language norms for studying bias in taboo detection.

\noindent 2. We use our method to assess bias against five minority communities in three taboo text classifiers and ten taboo annotated datasets.
In contrast, prior research has largely about bias against the African American community, possibly due to reliance on two race estimated/labelled datasets 
    \cite{blodgett-etal-2016-demographic, preotiuc-pietro-ungar-2018-user}.

\noindent 3. We find, for example, that all three taboo classifiers are biased against the African American community. Additionally, 2 classifiers are biased against the South Asian communities and one against Hispanics. Overall, only 3 of the 15 classifier - community combinations tested were somewhat unbiased. 
    
\noindent 4. Eight out of 10 taboo annotated test sets are strongly biased against African Americans, while at least 3 are biased against South Asians. We did not find any remarkable dataset biases against Hispanics but we found single instances of bias against Native American and Hawaiian  communities.

\section{Community Centered Approach for Studying Bias}

We build community-specific classification models (CLCs) which capture the community's language.
We use these to compute text alignment scores for assessing bias.
A text alignment score is the classifier's confidence in deciding if the text has been generated by a member of the community or not. Higher confidence scores imply greater alignment with the language norm's of the community. 

Our intuition is as follows.
Given moderation and self-regulation in communities we expect taboo utterances to be \textit{infrequent} in a community's discourse.
Such utterances will thus have low alignment scores when classified by a model trained on the community's language.
However, since new topics of interest will also result in low-frequency utterances 
we view low alignment scores with a community model as \emph{necessary} but \emph{insufficient} for an utterance to be considered taboo from that community's perspective.
Crucially, utterances with high alignment scores adhere to the norms and therefore should not be considered taboo. 
Reference classifier models allow us to estimate the extent of alignment with community language norms.

Two points to note. We do not attempt to separate style from topic in the utterances since it is often in their interplay that taboo is decided.  For example, \emph{``happy birthday to a bad bitch''} and \emph{``Fuck off, you are a little bitch.''} have somewhat related styles but different topics.
Our `contextual' community-language classification models estimate scores for full utterances.
Second, while we study race/ethnicity based communities, our approach can be applied to communities defined broadly, using other criteria such as professional communities and  internet subcultures. Specific details for measuring bias are given next.\\
\noindent
\textbf{Measuring bias in taboo classifiers:}
We assess the extent of taboo classifier bias against a community by  computing Pearson correlation between the taboo classifier confidence scores and community classifier confidence scores.
We do this with instances that the former classifier declares as taboo.
Ideally, we expect a negative correlation - higher taboo classifier confidence mapping to lower community alignment scores and vice versa.
The extent to which correlations deviate from this expectation reflects the extent to which the classifier does not consider the norms of the community.\\
%
%
\noindent
\textbf{Measuring bias in taboo datasets:}
%
Given a dataset, we compute the proportion of taboo labelled texts that are  highly aligned with each community classifier
model.
We expect these proportions to be tending towards zero since high alignment means common utterance and thus within the norms.
A second point to note is that if these proportions are uneven across communities then the bias is more against those with larger proportions.
This is consistent with the theory that all communities engage in similar norms of communication \cite{MILLS20091047}.
%
%
%
%
%
Specifically, we measure the mean and standard deviations (SD) of proportion of aligned comments.
In sum, if the proportions are not close to zero this indicates bias.
Additionally, if the proportions are not even then bias is targeting some communities more than others.
%

\section{Community Language Classifiers}

\begin{table}[]
    \centering
    \footnotesize
    \begin{tabular}{c|c|c|c}
                    &No. of&   Training & Validation \\
         Community& Subreddits & set size & set size\\\hline\hline
         NA & 2 & 44k & 1.4k\\\hline
         HI & 4 &95k & 6k\\\hline
        HA & 1 & 80k & 2k\\\hline
        SA & 1 &101k & 6k\\\hline
        AA & 11  & 70k & 5k\\\hline
    \end{tabular}
    \caption{Dataset details for each  community.}
    \label{tab:communities}
\end{table}


\subsection{Model Construction}
We build our community classification models using BERT 
\cite{devlin-etal-2019-bert}\footnote{Our code and data are available at: https://github.com/JonRusert/SuumCuique}. Experiments using XLNet \cite{yang2019xlnet} gave comparable results. Thus, we only report BERT results. BERT is a transformer-based model which has shown to perform well in NLP based tasks \cite{devlin-etal-2019-bert, 9058044}. BERT leverages context (on top of words)\footnote{For example, while ``is that all you have to contribute'' and ``contribute this have that is to you all'' have the same set of words, BERT will find the former more likely.}. We fine-tune BERT-base-uncased
\footnote{Models were trained on GeForce GTX 1080 Ti's, and took at most 3 hours to train. Parameter configuration is that of base-BERT, we use a dropout of 0.3 before the linear layer.}  with a linear layer on top. A softmax function is used to make a binary classification as to whether or not the input text belongs to the community. Each community has its own classifier representing its language norms. 

We build our models with publicly available data from select subreddit comments obtained using Pushshift \cite{baumgartner2020pushshift}\footnote{In our experiments, we do not use any data beyond that which is available on Pushshift. Pushshift is a public Reddit data repository.}.
We group subreddits into communities based on shared cultural/ethnic heritage
determined using subreddit descriptions (subreddits are listed in appendix A). While we acknowledge that subreddits are not inclusive of all members of communities of interest, they can still be considered as fairly representative in terms of language use.
We do not have to be exact about the data used to represent a community
as long as we are confident that the collection is mostly produced by its members\footnote{We also acknowledge that Reddit itself has some language norms. However, these norms should apply across all of our community subreddit datasets and so should not influence our results in significantly.}.

%
We build models for: Native American (NA), Hispanic (HI), Hawaii (HA), South Asian\footnote{South Asian English is considered non-standard compared to mainstream English.}
(SA), and African American (AA) communities.
We obtained comments from 2018 using the first 11 months as training data for the models and the last as validation. 
As the data for NA was smaller than the rest, we added 2015, 2016, and 2017 data again with the first 11 months for each year as training and last 1 month for validation. 
%
Comments were lowercased and stripped of punctuations. 

Since subreddit sizes vary, the amount of data collected for each community varies as well. HI, HA, SA, and AA are the closest in size and NA the smallest. 
%
%
When training a community model we generate negative samples (texts not from the community)
by sampling  equally from the other communities till we reach a 1:1 ratio of positive to negative text samples.
%
%
%
%
A summary of the datasets is in Table
\ref{tab:communities}.
Note that the training set size indicated includes both positive and negatives examples (at a 1:1 ratio) while the validation sets only contain positive (aligned) examples. 

\vspace{-0.5em}
\subsection{Model Characteristics} \label{sect:characteristics}
\vspace{-0.5em}
\begin{figure}
    \centering
    \includegraphics[width=1\columnwidth]{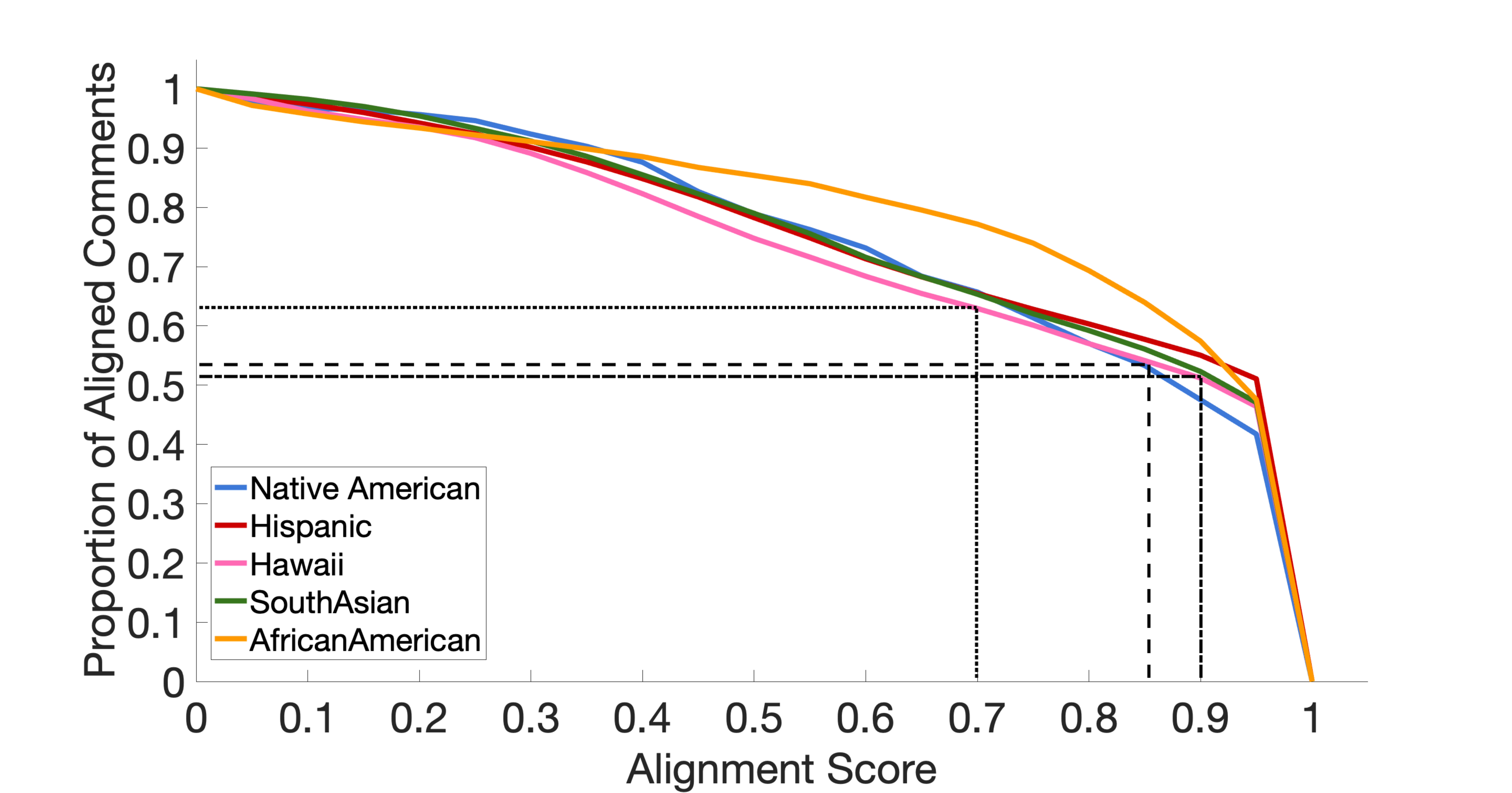}
    \caption{Complementary cumulative distribution function of community-language alignment scores for the five community models using the community specific validation datasets. An alignment score indicates the proportion of the community's comments that the model classifies as belonging to that community within a given threshold. The dotted/dotted-dashed lines (0.7, 0.9) show the proportion of comments at those thresholds. 
    The dashed line (0.85) indicates the chosen threshold for our experiments.}
    \label{fig:correspondingLM}
\end{figure}

\begin{figure}[]
    \centering
    \includegraphics[width=1\columnwidth]{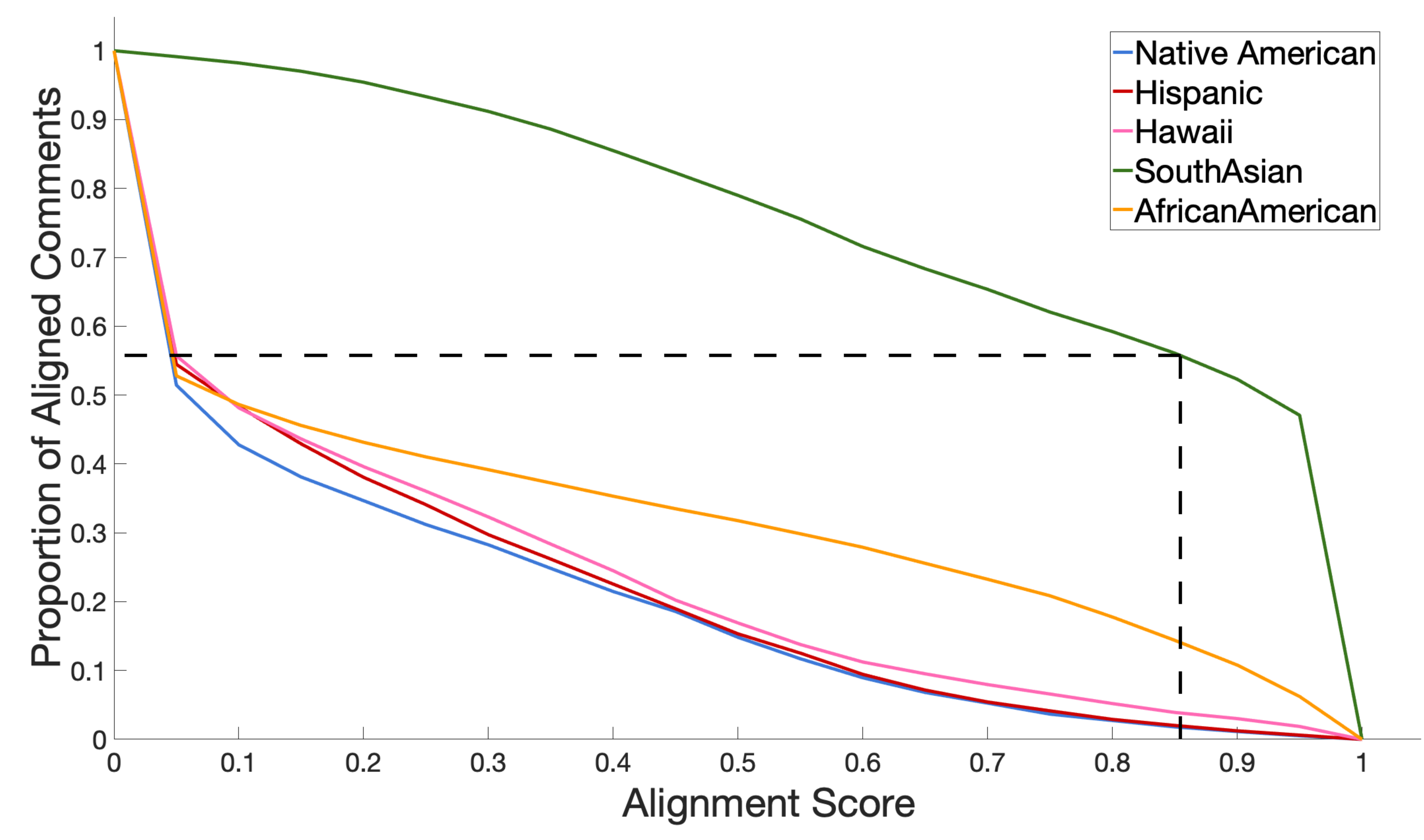}
    \caption{Complementary cumulative distribution function of alignment scores for all five minority community validation sets as gauged against the South Asian community's CLC. }
    \label{fig:correspondingSA}
\end{figure}

Figure \ref{fig:correspondingLM} presents the complementary cumulative distribution function (CCDF) of alignment scores (classifier confidence scores) for the CLCs using validation data. 
For example, 62.9\% and 66.2\% of the HA and the HI texts have scores $\geq$ 0.7.
All CCDFs follow a similar distributional pattern.
We introduce a threshold on alignment scores to decide which texts are `highly aligned' with a model.
Higher thresholds imply fewer comments will be regarded as following the community's norms.
We choose 0.85 as threshold since at least 52\% of the comments for each community are then regarded as highly aligned (range: 52\% - 64\%). 
When we reduce the threshold to 0.65,  two-thirds of the comments are highly aligned. But this increases the risk of overlap across communities in their highly aligned comments. 
%
Taking the SA CLC model as an example in Figure \ref{fig:correspondingSA}, we can see that with 0.85 cutoff, $60.7\%$ of comments from the South Asian community are highly aligned, while including no more than $14\%$ (range: 2\% - 14\%) of comments from the other communities.
While not strictly comparable, our 0.85 threshold is more conservative than the 0.80  used in \citet{blodgett-etal-2016-demographic}. 


\subsection{Model Validation} \label{sect:verification}
\vspace{-0.5em}
Results with our validation datasets (shown in Table \ref{tab:communities}) are in Table \ref{tab:LMTestSet} with cell values representing proportions.
For example, 51.8\% of the NA validation set is highly aligned (alignment score $\geq0.85$) with its own community model.
Column values do not necessarily add up to 100\% as a text may be highly aligned to more than one community model or even to none.
%

%

As expected, proportions at the diagonal representing homogenous model - dataset combinations  are high, ranging from 52 to 64\%. 
%
%
%
Also as expected, the off-diagonal entries representing heterogeneous combinations,   are low. Most are less than 10\% and more than half less than 6\%. The three noticeable exceptions are between AA and SA and also between NA texts and the HA model. 
The sociological literature
observes linguistic exchanges between minority communities  \cite{bucholtz1999borrowed,coleman1998language,igoudin2011asian,lee2011globalization}.
For example, 
\cite{shrikant2015yo}
discuss a tendency of South Asians to adopt the features of African American Vernacular English (AAVE).
This is particularly relevant to the AA and SA overlaps. Next we use these validated models to estimate bias in classifiers and in datasets.
\begin{table}[h]
    \centering
    \small
    \begin{tabular}{c||ccccc}
        & \multicolumn{5}{|c}{Reddit Validation  Sets} \\\hline
         CLC & NA & HI & HA & SA & AA\\\hline
         NA & \textbf{51.8} & 1.8 & 4.5 & 1.8 & 2.2  \\
         HI & 4.3 & \textbf{58.2} & 2.1 & 2.3  & 2.2  \\
         HA & 15.1 & 6.2 & \textbf{58.1} & 5.1  & 6.9  \\
         SA & 6.1 & 5.2 & 5.8 & \textbf{60.7} & 20.7  \\
         AA & 9.8 & 7.1 & 8.1  & 14.4 & \textbf{64.0}  \\  
         \hline
    \end{tabular}
    \caption{Proportion of each validation set that is highly aligned with each CLC. An alignment score threshold of 0.85 is used to determine high alignment. A text may be aligned with 0 or more models, so column numbers need not sum to 100.}
    \label{tab:LMTestSet}
\end{table}


\section{Experiments and Results}
\subsection{Taboo Classifier Bias Assessment}

We test 3 SOTA offensive language classifiers:


\noindent \textbf{NULI} \cite{liu-etal-2019-nuli}: A BERT \cite{devlin-etal-2019-bert} based system trained on offensive language (OLID \cite{zampierietal2019OLID}). It was the top-ranked system in OffensEval \cite{zampieri2019semeval}. 

\noindent \textbf{MIDAS} \cite{Mahata2019MIDASAS}: An ensemble of three deep learning models: a BLSTM, a BLSTM fed into a BGRU, and a CNN all three trained on offensive language. This system was the top non-BERT based system at OffensEval. 

\noindent \textbf{Perspective} \cite{Perspective}: An API provided by Google, which when given text, returns a toxicity score. The current model in production uses a CNN trained with fine-tuned GloVe word embeddings.

We trained MIDAS and NULI on the OLID training data which is consistent with their training for OffensEval.
Our implementations perform within 1\% of the published results. 
We then applied the classifiers to the validation datasets listed in Table \ref{tab:communities}.
Correlations between classifier confidence (for Perspective we use its toxicity score) and language model alignment scores were analyzed.

\subsection{Taboo Classifier Bias Results}
\begin{figure}
    \centering
    \includegraphics[width=0.75\columnwidth]{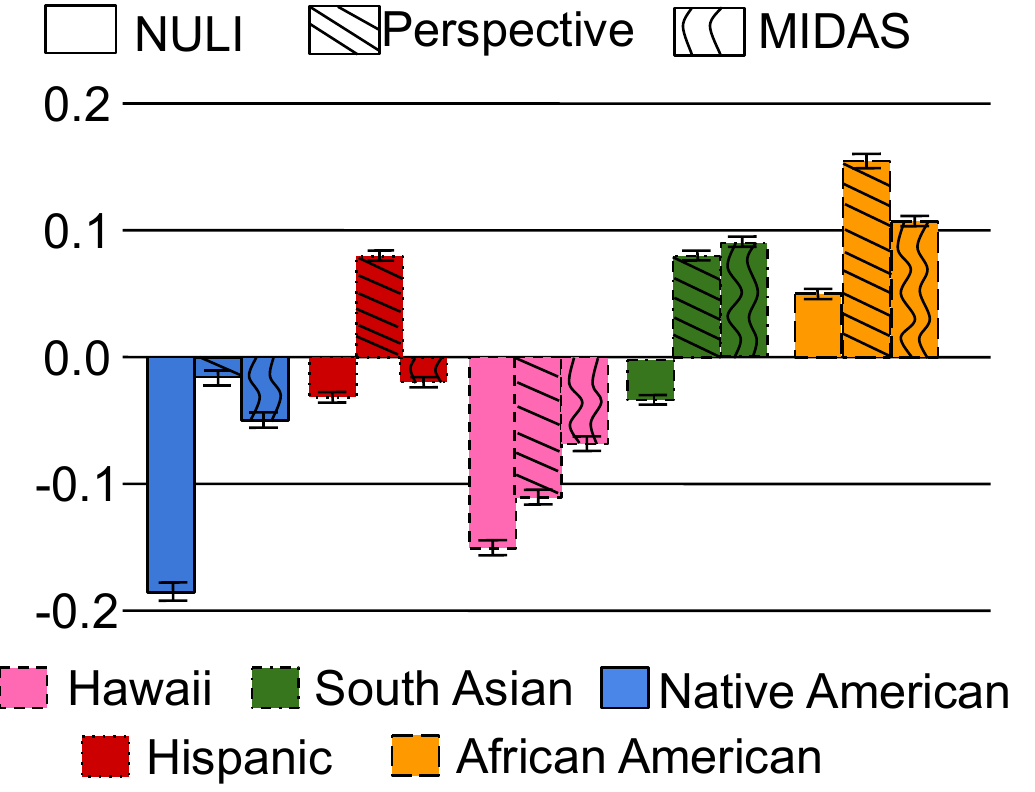}
    \caption{Correlations of taboo classifier scores with community-language classifier scores. Error bars: 95\% confidence intervals.}
    \label{fig:classifier}
\end{figure}

\noindent \textit{\textbf{Classifiers' correlations show bias.}}
Figure \ref{fig:classifier} shows the correlations with community models for the five communities for different taboo classifiers with
95\% confidence intervals computed using 10,000
bootstrapped samples.
Instead of strong negative correlations, which would indicate no bias, we see several instances of positive or near zero correlations across community models and classifiers. Zero correlations while better than positive are still not ideal, since they indicate that the taboo classifier is not taking community alignment into account.

NULI is relatively less biased than Perspective or MIDAS.
These correlations indicate that the taboo speech classifiers largely do not consider the language norms of the community. 

\noindent \textit{\textbf{Taboo classifiers have highest bias against the African American community.}} 
All three taboo classifiers tested show higher positive correlations for AA compared with other communities. 
This disagreement with the AA community model reflects bias in the classifiers, an observation that is consistent with
those made in prior  literature
\cite{davidson-etal-2019-racial,sap2019risk}.

\noindent \textit{\textbf{Perspective shows highest bias.}}
On average, Perspective, is the least in accordance - having the most positive correlations - with the CLC models. When compared with MIDAS and NULI, the only community model for which Perspective has a lower bias is HA. This is particularly concerning as Perspective is publicly available and has already been deployed to monitor comment sections \cite{HowElPa30:online,NYTimesPersp:online}.


\begin{table}[]
    \centering
    \footnotesize
    \begin{tabular}{c|c|c|p{1.5cm}}
         Dataset & Labels & Sizes & Prior Work \\\hline\hline
         Davidson & Offense & 20716 & Sap \citeyearpar{sap2019risk}\\
         \citeyearpar{davidson2017automated} & Hate & 1537 & Davidson \citeyearpar{davidson-etal-2019-racial}\\\hline
         OLID \citeyearpar{zampierietal2019OLID} & Offense & 4640 & None\\\hline
         SOLID \citeyearpar{rosenthal2020largeSOLID} & Offense & 3002 & None\\\hline
         Gab \citeyearpar{kennedy2018gab} & Hate & 2337 & Jin \citeyearpar{jin2020efficiently} \\\hline
         Founta & Hate & 27150 & Sap \\
         \citeyearpar{founta2018large} & Abuse & 4965 & \citeyearpar{sap2019risk} \\\hline
         \multirow{2}{*}{Wiki Toxic}\tablefootnote{\url{kaggle.com/c/jigsaw-toxic-comment\\ -classification-challenge}} & Toxic & 15295 & Vaidya \\
         & Hate & 1405 &\citeyearpar{vaidya2020empirical} \\\hline
         Waseem \citeyearpar{waseem-hovy-2016-hateful}& Sexist & 2673 & Davidson \citeyearpar{davidson-etal-2019-racial}\\\hline
    \end{tabular}
    \caption{Details of examined datasets. Samples of prior work which also examined these sets for bias are indicated in the final column.}
    \label{tab:datasets}
\end{table}

%

%
%
%

\begin{table*}[]
    \centering
    \footnotesize
    \begin{tabular}{c||c|c||c|c||c||c|c||c|c||c}
            & \multicolumn{2}{|c||}{Davidson} & & & Gab & \multicolumn{2}{|c||}{Founta} & \multicolumn{2}{|c||}{Wiki Toxic} & Waseem \\\hline
         CLC & HATE & OFF & OLID & SOLID  & Hate & Hate & Abuse & Toxic & Hate & Sexism\\\hline
         NA & 14.0 & 3.9 & 3.4 & 1.4 & 7.6 & 4.4 & 3.9 & 8.0 & \textbf{13.3} & 5.1\\
         HI & 5.5 & 5.2 & 8.3 & 3.5 & 5.5 & 5.4 & 6.6 & 4.9 & 6.3 & 3.9\\  
         HA & 4.3 & 3.1 & 6.3 & 5.1 & 3.9 & 6.0 & 4.6 & \textbf{9.4} & 3.4 & 4.2\\
         SA & 4.2 & 2.2 & \textbf{16.3} & 5.8 & \textbf{25.4} & 14.5 & 5.4 & 8.5 & \textbf{13.0} & 13.9\\
         AA & \textbf{20.7} & \textbf{29.9} & 15.2 & \textbf{30.4} & 12.2 & \textbf{32.6} & \textbf{22.5} & 4.9 & 5.3 & \textbf{45.5}\\\hline
         
         Average& 9.7 & 8.9 & 9.9 & 9.2 & 13.2 & 12.6 & 8.6 & 7.1 & 8.3 & 14.5\\\hline
         Std. Dev.&  7.4 & 11.8 & 5.6 & 11.9 & 8.7 & 11.9 & 7.8 & 2.1 & 4.6 & 17.8\\\hline
    \end{tabular}
    \caption{Proportion of  Taboo datasets with high alignment scores for each CLC. Note, a given text may have high alignment with 0 or more communities. Thus column proportions need not sum to 100.}
    \label{tab:datasetBias}
\end{table*}

\subsection{Dataset Bias Assessments and Results}
We generate ten test sets from seven text collections annotated for taboo. A test set consists of instances labeled with one taboo label (such as hateful, offensive, sexist etc.). A summary of the datasets can be found in Table \ref{tab:datasets}.

Table \ref{tab:datasetBias} presents bias assessment results.
Each row identifies the CLC used to estimate alignment of taboo instances in the column datasets.
For example, 14\% of the texts labelled as HATE in the Davidson  dataset are highly aligned with the Native American model.
We remind the reader that we use a community classifier score threshold of 0.85 (see section \ref{sect:characteristics}). Again column sums need not be 100 as an instance may be highly aligned with 0 or more models.
The table also shows average and standard deviation. 
Cells in bold are more than one standard deviation from the mean.

As a reminder, for any community we would like the proportion of taboo labeled instances that are highly aligned with that community's language to be close to 0. 
Additionally, if this is not possible, we check to see if these proportions are unequal across communities.
Inequality would indicate that bias is particularly targeted at some communities - those with highest proportions. 

\subsubsection{Taboo text and model alignments}
\noindent \textit{\textbf{Many of the proportions are large indicating bias.}}
When examining Table \ref{tab:datasetBias}, we find that 64\% of the cells have proportions $>$ 5\%. Even with a stronger threshold of $>$ 10\% on the proportions, close to a third of the cells still exhibit bias.
The least biased -- closest proportion to 0, is 1.4 for SOLID gauged against the NA model. The highest is 45.5 for Waseem gauged against the AA model.
Examining each dataset, we see that all have large proportions of taboo text that are highly aligned with at least one community; some datasets are biased against two communities.
For example, Wiki Toxic - Hate has high proportions when gauged against both NA and SA.
When examining averages across datasets Waseem is the most biased (average of 14.5), followed next by Gab.
Wiki Toxic - Toxic is the least biased --- with the lowest average.
%
Overall, all datasets  exhibit biases with proportions that are far from 0.

\subsubsection{Uneveness of alignment proportions}

\noindent\textit{\textbf{Datasets are heavily biased against African Americans.}}
As seen in table \ref{tab:datasetBias}, six out of the ten datasets have disproportionate high-alignment with the AA CLC model with a seventh (OLID) coming close to crossing the 1 SD mark.
The two datasets with lower than average proportions are from the Wiki Toxic collection.
The highest proportions are with the Waseem dataset 
followed by the Founta Hate dataset.
Telling too is that the top six in bias have more than 20\% of their instances highly aligned with the AA model.
This level and consistency of bias against the AA model is remarkable. 

\noindent\textit{\textbf{Datasets also biased against South Asians.}} Next to the AA community, the largest bias is against the SA community.
This community takes the top position with Gab Hate;  over a fourth of the data is highly aligned with the  community (25.4\%). 
Looking down the column this proportion for SA stands out, with the next highest --- AA, at 12.2\%.
In OLID and Wiki Toxic Hate, SA proportions are more than 1 SD from average.
SA alignments are lower than average for 5 datasets such as Davidson and 
Founta Abuse.
This analysis brings to light that bias is not limited to the AA community, but extends to other communities as well.
This is noteworthy as previous works such as \cite{davidson2017automated,sap2019risk} have solely focused on the bias against AA while ignoring others like SA.

\noindent\textit{\textbf{Low bias communities exist.}}
The HI and HA communities face less bias in comparison with AA and SA.
%
%
In fact, for HI the proportions are lower than average across all datasets.
%
NA has proportions 1 SD higher than average for one dataset: Wiki Toxic - Hate. 
Otherwise, the alignment proportions are almost always lower than average.

\noindent\textit{\textbf{Results are largely comparable with previous bias results.}}
Comparisons are limited as prior work largely focuses on bias towards AA.
We make the following general observations.

Our results are generally consistent with those of \citet{davidson-etal-2019-racial} 
and \citet{sap2019risk} on the Davidson dataset. 
E.g., all three find high rates of AA tweets as labelled offensive (between 17\% and 46\%).
We observe similar agreements for the Founta and Waseem sets.
There exists some disagreement, however, where we find the ``hate'' labeled portion of Davidson and Founta datasets to show bias against AA, while \citet{sap2019risk} did not. 
The reasons for this difference are unclear.
However, it is puzzling to note that \citet{sap2019risk} found bias against a community for one taboo class (offense) but not another (hate). The authors do not provide an explanation for this.



\citet{huang-etal-2020-multilingual} 
examining the Waseem and Founta datasets for bias against race find BERT to show the least bias compared to CNN and RNN based classifiers. Our results are consistent as the BERT based classifier, NULI, shows lower bias than MIDAS, which is an ensemble of a CNN, and BLSTM/BGRUs. 

Overall, our results indicate that the datasets are largely insensitive to the norms of African Americans (overwhelmingly so) and then to the South Asian communities.
Subsequently, taboo classifiers built on these datasets are more likely to be biased against these two communities.

\subsection{Small scale user study}
We present a small scale user study intended only to illustrate our main hypothesis, that 
texts highly aligned with a CLC should be less likely to be considered taboo from the perspective of members of that community. 
A rigorous user study with intra and inter community judgements and many judges is left for future work.

\noindent
\textbf{{Classifier bias.}} Focusing on Figure \ref{fig:classifier}, we selected a mixture of comments from OLID/SOLID and from reddit that had the highest model alignment scores ($>0.85$) with their respective CLCs. We selected 80 comments for AA and 78 for SA, all of which had high taboo classifier confidence scores ($> 0.76$) with MIDAS and Perspective, i.e., they contributed the most to the positive correlation bars in the graphs. 
We asked two African Americans and two South Asians, who were active Reddit users, to judge their respective sets as offense/hate or not. The annotators were selected by word of mouth. 
Given high alignment we expect annotators to contradict the classifier and judge these as not taboo.

As expected, both SA annotators disagreed with the classifier assigned taboo labels in 60/78 cases (76.9\%) agreeing only in 3/78 comments (3.8\%) and giving mixed judgements in the remaining 19\% of comments.
AA  annotators
disagreed with the classifiers for 27/80 (33.8\%) comments. While non trivial, this percentage is noticeably less than 76.9\% for SA. They agreed with the classifier's taboo decision 31/80 times (38.8\%) and gave mixed judgements in 25\% of cases (more than for AA).  
Overall, judges contradicted the classifiers 55\% of the time.

\noindent\textbf{Analysis of classifier contradictions.}
In a majority of the contradictions the  classifier did not recognize benign contexts of words such as `fuck' and `shit'. E.g., in SA are: \textit{``how the fuck are Indians minorities'} and \textit{`how is an accent shitty''}. From AA: 
\textit{``... drop this fake ass bitch and move on''}.
Annotators did not label these as offensive.

A second reason for contradictions appears to be because culturally subtle contexts are challenging for classifiers. 
Both SA annotators marked the comment \textit{``this is exactly the reason i don't fuck with Biryani"} as non-taboo. The classifier does not recognize this as a statement on the authenticity of a food item -   `Biryani', a rice-based South Asian dish.  
Similarly the AA annotators marked:  \textit{``I'm so fucking tired of side chick culture''} as not taboo.  The classifier does not recognize this as a critique of  extra-marital intimacy and not an attack on a `culture'. Note that all these statements have model alignment scores that are very high ($> 0.94$).

\noindent\textbf{Dataset Bias.}
Focusing on 
OLID (biased against South Asians) and SOLID (biased against African Americans), see table \ref{tab:communities}, we explore if community consistent judges will contradict dataset annotations.
We selected 
50 SOLID and 23 OLID tweets annotated in the datasets as taboo and with CLC model alignment score $\geq 0.85$.


Both AA annotators contradicted taboo annotations for 66\% (30/50) SOLID tweets and  agreed with only 9/50 (18\%).  They mutually disagreed on the remaining 11 tweets.
Both SA annotators contradicted  83\% (19 of 23) of OLID taboo annotations. They mutually disagreed on the remaining 4.
Overall 67\% of community-specific annotator decisions contradicted taboo annotations. These contradictions are also illustrative of  our expectations of `not taboo' decisions for high alignment posts.

\noindent\textbf{Analysis of annotation contradictions.}
AA annotators chose `not taboo' in several SOLID cases based on context and their norms.  E.g., \textit{``this bitch just cut her hair short by herself''}. Perhaps the presence of language likely considered offensive in a more general setting  resulted in the taboo label.
In another example, the AA annotators contradicted the SOLID taboo label for the tweet \textit{``Niggas be so depressed on this lil app ...''}. 
We see a similar phenomenon with the SA annotators. For example, they contradicted the taboo label for the tweet \textit{``All these sick ass ppl from school gave me something and now I have to chug down this nasty drink so it can go away''}. 

Additional examples of contradictions by AA and SA annotators are in the appendix (B and C).
While a larger user-level study with suitable controls is planned for the future, these preliminary results illustrate our hypothesis that using a community perspective on taboo decisions is important and that this can be achieved using CLC model alignment scores.

\subsection{Bias mitigation - initial thoughts}

Our empirical results and case study illustrate a potential approach for bias mitigation.
The idea is to include `reset' strategies for taboo classifiers and for annotations.
If an instance that is classified or annotated as taboo has high alignment score with a CLC then a reset strategy is invoked to examine the  decision further.
This can be manual analysis by a community member knowledgeable of its norms and contexts - a step that can be labour expensive.
Alternatively, a downstream algorithm that is more community centric may be invoked - a step that may computationally more expensive. 
We plan to explore these in future.




\section{Related Work}

The study of bias in taboo classifiers is an active area of research \cite{wiegand2019detection, huang-etal-2020-multilingual,blodgett-etal-2020-language, park-etal-2018-reducing}.  
Since our focus is on bias detection we do not cover areas such as bias mitigation \cite{tsvetkov2020demoting,chuang2021mitigating} and allied problems such as bias in embedding spaces \cite{park-etal-2018-reducing,Caliskan183} and the source of bias  \cite{binns2017like,waseem2016you}.
Instead we focus on bias detection - particularly on their methodologies.
Note as most papers combine methods our binning is not intended as mutually exclusive classes.




\noindent \textbf{Bias detection focused on individual words:}
Dixon et al. \citeyearpar{dixon2018measuring} working with Wikipedia Talk Pages show that toxicity classifiers disproportionately misclassify text with identity terms such as `gay' and `muslim' when they appear in benign contexts. 
 A reason for this bias is because of their dominant occurrence in hateful, toxic or otherwise taboo contexts, thereby skewing training datasets. 
 They show this can be countered by augmenting data with benign examples using these identity words. 
\citet{park-etal-2018-reducing}, and \citet{kennedy2018gab} also study similar biases in the data.
Badjatiya et al., \citeyearpar{badjatiya2019stereotypical} extend this analysis to words in general, finding strings like `@ABC' and `dirty' stereotypical of hate.
The general approach in these papers is to compare classifiers built from skewed and synthetically augmented/corrected non-skewed datasets.
In contrast to focusing on specific words, we analyze texts contextually with CLCs. 

\noindent \textbf{Bias detection using an external race/gender labeled dataset:}
Davidson et al. \citeyearpar{davidson-etal-2019-racial} examine five datasets including the Davidson and Founta datasets we study
. 
Their strategy is to build a classifier (regularlized logistic regressor) on each dataset and test them on the \citet{blodgett-etal-2016-demographic} black-white-race-aligned dataset.
They observed that `black-aligned' tweets were 1.8 to 2.6 times more likely to be classified as taboo.
Using similar methodology, Sap et al. \citeyearpar{sap2019risk} also found bias against AA in neural network classifiers trained on the Founta and Davidson datasets  
when tested against the same \citet{blodgett-etal-2016-demographic} dataset.


Using a slightly different approach \citet{kim2020intersectional} trained a classifier on the \citet{blodgett-etal-2016-demographic} dataset to identify AA-leaning tweets in the Founta dataset. 
They found that AA tweets were 3.7 times more likely to be labeled as taboo.
Additionally, they also annotated the Founta dataset for gender. 
They found that AA male-aligned tweets were 77\% more likely to be labeled as taboo. 
In contrast to these works, our approach does not rely on access to race labeled datasets such as the \citet{blodgett-etal-2016-demographic} dataset.



\noindent\textbf{Other approaches:}
As in \citet{kim2020intersectional},
\citet{sap2019risk} used classifiers trained on  \citet{blodgett-etal-2016-demographic} to identify `black-aligned' tweets in
the Founta and Davidson datasets. 
The difference is that they then computed correlation of probabilities of `black-aligned' tweets and the taboo language annotations and found these to be strong positive. 
We also compute Pearson correlations between classifier scores and community alignment scores. But we differ in that we are not relying on the \citet{blodgett-etal-2016-demographic} dataset.  Instead, we use alignment scores estimated using CLCs.

While the above works highlight limitations in using a single standard English model they do not propose alternative methods founded on the language norms of specific communities. We address this gap with a community-specific method using CLCs which allows us to study bias for any community. 
Moreover, unlike the works reviewed, and more recent papers on debiasing strategies (\citet{zhou2021challenges,xu2021detoxifying}), where the emphasis is on bias against AA, we study bias against five different minority communities.
\section{Limitations and Conclusions}

%
%
We presented a new methodology for studying bias in taboo text  identification.
Its strength is that it is centred on community language norms - a strategy consistent with \cite{blodgett-etal-2020-language} to consider social hierarchy when studying bias and natural languages.
Using it we assessed the extent to which biases against
five minority communities are present - both in classifier taboo decisions and in dataset taboo annotations.
We found many instances of  bias with
the community most targeted being African Americans.  
But we also found significant biases against others such as South Asians.
Notably, Hispanics seems least affected though there was to some degree classifier bias.
%
%

 
%
%
A small scale `illustrative' user study provides initial support for our key idea which is that common utterances, i.e., those with high alignment with the community's language classifier, are unlikely to be viewed as taboo from that community's perspective.
Annotators who are community members contradicted classifier and annotator decisions in a majority of instances.
%

%
%

Our work is limited to analyzing communities defined by race and ethnicity, but is generalizable; we will explore bias against other  communities in future research.
We also plan to conduct a large-scale user study to better understand community perspective for taboo texts.
We plan to 
investigate the bias mitigation strategies shown in section 4.5.
%




\bibliography{main}
\bibliographystyle{acl_natbib}

\appendix
\appendix
\onecolumn
\section{Community Subreddits}
\begin{table}[h!]
    \centering
    \begin{tabular}{|c|p{6cm}|}\hline
         \textbf{Community} & \textbf{Subreddits} \\\hline\hline
         Native American &  r/IndianCountry,                    r/NativeAmerican \\\hline
         Hispanic & r/LatinoPeopleTwitter, 
                    r/Chicano,
                    r/cuba \\\hline
        Hawaii &  r/Hawaii \\\hline
        South Asian & r/ABCDesis \\\hline
        African American & r/AfricanAmerican,                    r/asklatinamerica,
                           r/blackladies,
                           r/blackfellas,
                           r/blacklivesmatter,
                           r/blackcreatives,
                           r/blackhair,
                           r/blackwomens,
                           r/ebonyimagination,
                           r/blackcontemporaryart,
                           r/afrocentrism,
                           r/afrofuturism \\\hline
    \end{tabular}
    \caption{Communities studied and their corresponding subreddits.}
    \label{tab:subreddits}
\end{table}

\section{Guideline for user study}

WARNING: The following comments may contain  offensive/hateful material.  These might includes instances of racism, sexism and misogyny.  If at any point you feel you cannot judge any more  then just send us what you complete.
Thank you for your support of our research

\vspace{1em}

\noindent TASK description:

\noindent Based on the language norms used in your cultural/ethnic/racial background, do you find the following texts offensive or hateful? Indicate your decision with a 'Yes' in column A for texts you feel are offensive/hateful and 'No' otherwise.

\newpage
\section{African American examples from the user study}
\begin{table}[h!]
    \centering
    \begin{tabular}{|p{8cm}|l|l|l|}
    \hline
    \textbf{Text}&\textbf{CLC Score}&\textbf{MIDAS Score}& \textbf{Persp. Score}\\\hline
   aw damn im so sorry sis that is the worst&0.99& 0.96 & 0.78\\\hline
 im so fucking tired of side chick culture&0.99& 0.96 & 0.90\\\hline
 lol i caught so much shit with this so i cant even imagine if you brought out colorism&0.99& 0.95 & 0.86\\\hline
 i know who it might be but i wont say because i dont want to get banned for starting shit&0.98& 0.94 & 0.90\\\hline
 smh these niggas aint shit lol my now ex boyfriend is almost 30 and all he do is play nintendo all day and doesnt wanna seek a job smh&0.99& 0.93 & 0.95\\\hline
 i would have ghosted his ass so fast&0.96& 0.91 & 0.94\\\hline


i need lasik eye surgery i'm blind as a bitch this shit can't be healthy&0.99& OFF & OFF\\\hline
 The older you get the more you  understand that vibez is more important than a cute face \& a fat ass&0.85&OFF & OFF\\\hline
 I never got to thank her for everything she has done for me as a Black woman \& a writer. I am so fucking sad this hurts so fucking much&0.99&OFF & OFF\\\hline
 My family is coming.. I need to put a bra on but damn I'm lazy&0.99&OFF & OFF\\\hline
 i turn into the craziest bitch every leo season&0.99&OFF & OFF\\\hline

\end{tabular}
    \caption{Examples of comments that the classifiers or annotators labeled taboo (offensive/toxic/hateful), but both African American annotators agreed were non-taboo. Note that MIDAS and Persp scores which indicate ``OFF'' are those labels of the respective original dataset.}
    \label{tab:classifier-AA}
\end{table}

\twocolumn
\newpage
\onecolumn
\section{South Asian examples from the user study}

\begin{table*}[h]
    \centering
    \begin{tabular}{|p{8cm}|l|l|l|}
    \hline
   \textbf{Text}&\textbf{CLC Score}&\textbf{MIDAS Score}& \textbf{Persp. Score}\\\hline

it is not unreasonable to assume she was with a desi guy calm your fucking tits& 0.99 & 0.97 & 0.93 \\\hline
 lol dil mil is crap anyway i get the same 10 girls on repeat and im like dafuq is wrong with this shit & 0.99 & 0.96 & 0.93\\\hline
 lol the idiot who thinks india has no discrimination is going to preach to me stop killing people for eating beef or even suspected of eating beef and come back to me you fucking hypocrite& 0.99 & 0.97 & 0.95\\\hline
 really darkskinned desi girls are fucking seeexxxy& 0.99 & 0.96 & 0.96\\\hline
 yeah no shit you dont even live in india& 0.99 & 0.96 & 0.93\\\hline
 exactly tribalism all ready divided and fucked up south asia why do we have to spill that shit over to the us too& 0.98 & 0.95 & 0.93\\\hline
 this is exactly the reason i dont fuck with biryani& 0.99 & 0.95 & 0.89\\\hline
 shitty parenting and cultural, religious customs will do that to you& 0.96 & 0.76 & 0.93\\\hline

\#ArunJaitleyStepDown He is most shameless \#FM in history of India and audacity and shamelessness with which is lies in public is disgrace to post.& 0.99 & OFF &OFF \\\hline
 \#FailedDemFeinstein should have quit decades ago. She is a disgrace.  Feinstein blames GOP after Kavanaugh accuser stays mum, admits 'I can't say everything's truthful'  URL & 0.85 & OFF &OFF \\\hline
 \#JusticeForSoniasFather  Mr Usman buzdar sb! please respond to Sonia Iqbal daughter of PTI counsellor from okara whose father was killed during elections by nawaz league killers. She is saying she will commit suicide in front of media. Where is justice? & 0.99 & OFF &OFF \\\hline

         \end{tabular}
    \caption{Examples of comments that the classifiers or annotators labeled taboo (offensive/toxic/hateful), but both South Asian annotators agreed were non-taboo. Note that MIDAS and Persp scores which indicate ``OFF'' are those labels of the respective original dataset.}
    \label{tab:classifier-AA}
\end{table*}

\end{document}